\pdfoutput=1

\documentclass[11pt,a4paper]{article}
\usepackage[hyperref]{acl2021}
\usepackage{times}
\usepackage{latexsym}

\usepackage{microtype}

\aclfinalcopy %
\def\aclpaperid{1426} %

\usepackage{amsmath,amsfonts,bm}
\usepackage{booktabs} %
\usepackage{url}
\usepackage{graphicx}
\usepackage{array}
\usepackage{color}
\usepackage{makecell}
\usepackage{multirow}
\usepackage{xspace}

\newcommand{\ie}{\textit{i.e.}} %
\newcommand{\eg}{\textit{e.g.}} %
\newcommand{\start}[1]{\vspace{.4mm}\noindent{{\bf #1}.}}

\newcommand{\upv}{\vspace{-.0cm}}
\newcommand{\downv}{\vspace{-.1cm}}
\newcommand{\upvf}{\vspace{-.2cm}}

\newcommand{\ours}{\textsc{Gar}\xspace}
\newcommand{\oursPlus}{$\textsc{Gar}^{\texttt{+}}$\xspace}

\definecolor{gred}{RGB}{219,68,55}
\definecolor{gblue}{RGB}{66,133,244}
\definecolor{gyellow}{RGB}{244,180,0}
\definecolor{ggreen}{RGB}{15,157,88}
\definecolor{ggrey}{RGB}{115,115,115}

\newcommand{\colorR}[1]{\textcolor{gred}{\textbf{#1}}}
\newcommand{\colorG}[1]{\textcolor{ggreen}{\textbf{#1}}}
\newcommand{\colorB}[1]{\textcolor{gblue}{#1}}

\title{Generation-Augmented Retrieval for Open-Domain Question Answering}

\author{\makecell{Yuning Mao$^{1}$\thanks{\hspace{.06in}Work was done during internship at Microsoft Azure AI.}, Pengcheng He$^{2}$, Xiaodong Liu$^{3}$, Yelong Shen$^2$,\\ Jianfeng Gao$^{3}$, Jiawei Han$^1$, Weizhu Chen$^2$} \\
$^1$University of Illinois, Urbana-Champaign \quad
$^2$Microsoft Azure AI \quad
$^3$Microsoft Research\\
$^1$\{yuningm2, hanj\}@illinois.edu \\ $^{2,3}$\{penhe, xiaodl, yeshe, jfgao,wzchen\}@microsoft.com 
} 

\date{}

\begin{document}
\maketitle
\begin{abstract}
We propose Generation-Augmented Retrieval (\ours) for answering open-domain questions, which augments a query through text generation of heuristically discovered relevant contexts without external resources as supervision. 
We demonstrate that the generated contexts substantially enrich the semantics of the queries and \ours with sparse representations (BM25) achieves comparable or better performance than state-of-the-art dense retrieval methods such as DPR~\cite{karpukhin2020dense}.
We show that generating diverse contexts for a query is beneficial as fusing their results consistently yields better retrieval accuracy.
Moreover, as sparse and dense representations are often complementary, \ours can be easily combined with DPR to achieve even better performance.
\ours achieves state-of-the-art performance on Natural Questions and TriviaQA datasets under the extractive QA setup when equipped with an extractive reader, and consistently outperforms other retrieval methods when the same generative reader is used.\footnote{Our code and retrieval results are available at \url{https://github.com/morningmoni/GAR}.}
\end{abstract}

\section{Introduction}
Open-domain question answering (OpenQA) aims to answer factoid questions without a pre-specified domain and has numerous real-world applications.
In OpenQA, a large collection of documents (\eg, Wikipedia) are often used to seek information pertaining to the questions.
One of the most common approaches uses a retriever-reader architecture \cite{chen-etal-2017-reading}, which first retrieves a small subset of documents \textit{using the question as the query} and then reads the retrieved documents to extract (or generate) an answer.
The retriever is crucial as it is infeasible to examine every piece of information in the entire document collection (\eg, millions of Wikipedia passages) and the retrieval accuracy bounds the performance of the (extractive) reader.

Early OpenQA systems \cite{chen-etal-2017-reading} use classic retrieval methods such as TF-IDF and BM25 with sparse representations. Sparse methods are lightweight and efficient, but unable to perform semantic matching and fail to retrieve relevant passages without lexical overlap.
More recently, methods based on dense representations \cite{guu2020realm,karpukhin2020dense} learn to embed queries and passages into a latent vector space, in which text similarity beyond lexical overlap can be measured.
Dense retrieval methods can retrieve semantically relevant but lexically different passages and often achieve better performance than sparse methods.
However, the dense models are more computationally expensive and suffer from information loss
as they condense the entire text sequence into a fixed-size vector
that does not guarantee exact matching \cite{luan2020sparse}.

 There have been some recent studies on query reformulation with text generation for other retrieval tasks, which, for example, rewrite the queries to context-independent~\cite{yu2020few,lin2020query,vakulenko2020question} or well-formed~\cite{liu2019generative} ones.
However, these methods require either task-specific data (\eg, conversational contexts, ill-formed queries) or external resources such as paraphrase data~\cite{zaiem2019sequence,wang2020deep} that cannot or do not transfer well to OpenQA.
Also, some rely on time-consuming training process like reinforcement learning (RL)~\cite{nogueira-cho-2017-task,liu2019generative,wang2020deep} that is not efficient enough for OpenQA (more discussions in Sec.~\ref{sec:related_work}).

In this paper, we propose Generation-Augmented Retrieval (\ours), which augments a query through text generation of a pre-trained language model (PLM).
Different from prior studies that reformulate queries, \ours does not require external resources or downstream feedback via RL as supervision, because it does not \textit{rewrite} the query but \textit{expands} it with heuristically discovered relevant contexts, which are fetched from PLMs and provide richer background information (Table~\ref{tab:generation_eg}). 
For example, by prompting a PLM to generate the title of a relevant passage given a query and appending the generated title to the query, it becomes easier to retrieve that relevant passage.
Intuitively, the generated contexts explicitly express the search intent not presented in the original query.
As a result, \ours with sparse representations achieves comparable or even better performance than state-of-the-art approaches~\cite{karpukhin2020dense,guu2020realm} with dense representations of the original queries, while being more lightweight and efficient in terms of both training and inference (including the cost of the generation model) (Sec.~\ref{sec:runtime}).

Specifically, we expand the query (question) by adding relevant contexts as follows.
We conduct seq2seq learning with the question as the input and various freely accessible in-domain contexts as the output such as \textit{the answer, the sentence where the answer belongs to}, and \textit{the title of a passage that contains the answer}. 
We then append the generated contexts to the question as the \textit{generation-augmented query} for retrieval.
We demonstrate that using multiple contexts from diverse generation targets is beneficial 
as fusing the retrieval results of different generation-augmented queries consistently yields better retrieval accuracy.

We conduct extensive experiments on the Natural Questions (NQ) \cite{kwiatkowski-etal-2019-natural} and TriviaQA (Trivia) \cite{joshi-etal-2017-triviaqa} datasets. 
The results reveal four major advantages of \ours:
(1) \ours, combined with BM25, achieves significant gains over the same BM25 model that uses the original queries or existing unsupervised query expansion (QE) methods.
(2) \ours with sparse representations (BM25) achieves comparable or even better performance than the current state-of-the-art retrieval methods, such as DPR \cite{karpukhin2020dense}, that use dense representations.
(3) Since \ours uses sparse representations to measure lexical overlap\footnote{Strictly speaking, \ours with sparse representations handles semantics before retrieval by enriching the queries, while maintaining the advantage of exact matching.}, it is complementary to dense representations: by fusing the retrieval results of \ours and DPR (denoted as \oursPlus), we obtain consistently better performance than either method used individually.
(4) \ours outperforms DPR in the end-to-end QA performance (EM) when the same extractive reader is used: EM=41.8 (43.8 for \oursPlus) on NQ and 62.7 on Trivia, creating new state-of-the-art results for extractive OpenQA. \ours also outperforms other retrieval methods under the generative setup when the same generative reader is used: EM=38.1 (45.3 for \oursPlus) on NQ and 62.2 on Trivia.

\start{Contributions}
(1) We propose Generation-Augmented Retrieval (\ours), which augments queries with heuristically discovered relevant contexts through text generation without external supervision or time-consuming downstream feedback. 
(2) We show that using generation-augmented queries achieves significantly better retrieval and QA results than using the original queries or existing unsupervised QE methods. %
(3) We show that \ours, combined with a simple BM25 model, achieves new state-of-the-art performance on two benchmark datasets in extractive OpenQA and competitive results in the generative setting.

\section{Related Work}
\label{sec:related_work}
\start{Conventional Query Expansion}
\ours shares some merits with query expansion (QE) methods based on pseudo relevance feedback \cite{rocchio1971relevance,abdul2004umass,lv2010positional} in that they both expand the queries with relevant contexts (terms) without the use of external supervision. \ours is superior as it expands the queries with knowledge stored in the PLMs rather than the retrieved passages and its expanded terms are learned through text generation.

\start{Recent Query Reformulation}
There are recent or concurrent studies \cite{nogueira-cho-2017-task,zaiem2019sequence,yu2020few,vakulenko2020question,lin2020query} that reformulate queries with generation models for other retrieval tasks.
However, these studies are not easily applicable or efficient enough for OpenQA because: (1) They require external resources such as paraphrase data~\cite{zaiem2019sequence}, search sessions~\cite{yu2020few}, or conversational contexts~\cite{lin2020query,vakulenko2020question} to form the reformulated queries, which are not available or showed inferior domain-transfer performance in OpenQA~\cite{zaiem2019sequence};  (2) They involve time-consuming training process such as RL. For example, \citet{nogueira-cho-2017-task} reported a training time of 8 to 10 days as it uses retrieval performance in the reward function and conducts retrieval at each iteration.
In contrast, \ours uses freely accessible in-domain contexts like passage titles as the generation targets and standard seq2seq learning, which, despite its simplicity, is not only more efficient but effective for OpenQA.

\start{Retrieval for OpenQA} 
Existing sparse retrieval methods for OpenQA \cite{chen-etal-2017-reading} solely rely on the information of the questions.
\ours extends to contexts relevant to the questions by extracting information inside PLMs and helps sparse methods achieve comparable or better performance than dense methods~\cite{guu2020realm,karpukhin2020dense}, while enjoying the simplicity and efficiency of sparse representations.
\ours can also be used with dense representations to seek for even better performance, which we leave as future work.

\start{Generative QA}
Generative QA generates answers through seq2seq learning instead of extracting answer spans.
Recent studies on generative OpenQA \cite{lewis2020retrieval,min2020ambigqa,izacard2020leveraging} are orthogonal to \ours in that they focus on improving the reading stage and directly reuse DPR \cite{karpukhin2020dense} as the retriever.  
Unlike generative QA, the goal of \ours is not to generate perfect answers to the questions but pertinent contexts that are helpful for retrieval.
Another line in generative QA learns to generate answers without relevant passages as the evidence but solely the question itself using PLMs \cite{roberts2020much,brown2020language}. 
\ours further confirms that one can extract factual knowledge from PLMs, which is not limited to the answers as in prior studies but also other relevant contexts.

\section{Generation-Augmented Retrieval}
\subsection{Task Formulation}
OpenQA aims to answer factoid questions without pre-specified domains. We assume that a large collection of documents $C$ (\ie, Wikipedia) are given as the resource to answer the questions and a retriever-reader architecture is used to tackle the task, where the retriever retrieves a small subset of the documents $D \subset C$ and the reader reads the documents $D$ to extract (or generate) an answer.
Our goal is to improve the effectiveness and efficiency of the retriever and consequently improve the performance of the reader. 

\subsection{Generation of Query Contexts}
In \ours, queries are augmented with various heuristically discovered relevant contexts in order to retrieve more relevant passages in terms of both quantity and quality.
For the task of OpenQA where the query is a question, we take the following three freely accessible contexts as the generation targets.
We show in Sec.~\ref{res_retrieval} that having multiple generation targets is helpful in that fusing their results consistently brings better retrieval accuracy.

\start{Context 1: The default target (answer)}
The default target is the label in the task of interest, which is the answer in OpenQA.
The answer to the question is apparently useful for the retrieval of relevant passages that contain the answer itself.
As shown in previous work \cite{roberts2020much,brown2020language}, PLMs are able to answer certain questions solely by taking the questions as input (\ie, closed-book QA).
Instead of using the generated answers directly as in closed-book QA, \ours treats them as contexts of the question for retrieval. The advantage is that even if the generated answers are partially correct (or even incorrect), they may still benefit retrieval as long as they are relevant to the passages that contain the correct answers (\eg, co-occur with the correct answers).

\start{Context 2: Sentence containing the default target}
The sentence in a passage that contains the answer is used as another generation target.
Similar to using answers as the generation target, the generated sentences are still beneficial for retrieving relevant passages even if they do not contain the answers, as their semantics is highly related to the questions/answers (examples in Sec.~\ref{sec:exp_gen}).
One can take the relevant sentences in the ground-truth passages (if any) or those in the positive passages of a retriever as the reference, depending on the trade-off between reference quality and diversity.

\start{Context 3: Title of passage containing the default target}
One can also use the titles of relevant passages as the generation target if available.
Specifically, we retrieve Wikipedia passages using BM25 with the question as the query, and take the page titles of positive passages that contain the answers as the generation target. 
We observe that the page titles of positive passages are often entity names of interest, and sometimes (but not always) the answers to the questions.
Intuitively, if \ours learns which Wikipedia pages the question is related to, the queries augmented by the generated titles would naturally have a better chance of retrieving those relevant passages.

While it is likely that some of the generated query contexts involve unfaithful or nonfactual information due to hallucination in text generation \cite{mao2020constrained} and introduce noise during retrieval, they are beneficial rather than harmful overall, as our experiments show that \ours improve both retrieval and QA performance over BM25 significantly.
Also, since we generate 3 different (complementary) query contexts and fuse their retrieval results, the distraction of hallucinated content is further alleviated.

\subsection{Retrieval with Generation-Augmented Queries}
\label{sec:retrieval}
After generating the contexts of a query, we append them to the query to form a \textit{generation-augmented query}.\footnote{One may create a title field during document indexing and conduct multi-field retrieval but here we append the titles to the questions as other query contexts for generalizability.}
We observe that conducting retrieval with the  generated contexts (\eg, answers) alone as queries instead of concatenation is ineffective because (1) some of the generated answers are rather irrelevant, and (2) a query consisting of the correct answer alone (without the question) may retrieve false positive passages with unrelated contexts that happen to contain the answer. Such low-quality passages may lead to potential issues in the following passage reading stage.

If there are multiple query contexts, we conduct retrieval using queries with different generated contexts separately and then fuse their results. The performance of one-time retrieval with all the contexts appended is slightly but not significantly worse.
For simplicity, we fuse the retrieval results in a straightforward way: an equal number of passages are taken from the top-retrieved passages of each source.
One may also use weighted or more sophisticated fusion strategies such as reciprocal rank fusion \cite{cormack2009reciprocal}, the results of which are slightly better according to our experiments.\footnote{We use the fusion tools at \url{https://github.com/joaopalotti/trectools}.}

Next, one can use any off-the-shelf retriever for passage retrieval.
Here, we use a simple BM25 model to demonstrate that \ours with sparse representations can already achieve comparable or better performance than state-of-the-art dense methods while being more lightweight and efficient (including the cost of the generation model), closing the gap between sparse and dense retrieval methods.

\section{OpenQA with \ours}
To further verify the effectiveness of \ours, we equip it with both extractive and generative readers for end-to-end QA evaluation. We follow the reader design of the major baselines for a fair comparison, while virtually any existing QA reader can be used with \ours.

\subsection{Extractive Reader}
For the extractive setup, we largely follow the design of the extractive reader in DPR \cite{karpukhin2020dense}.
Let $D = [d_1, d_2, ..., d_k]$ denote the list of retrieved passages with passage relevance scores $\mathbf{D}$. Let $S_i = [s_1, s_2, ..., s_N]$ denote the top $N$ text spans in passage $d_i$ ranked by span relevance scores $\mathbf{S_i}$.
Briefly, the DPR reader uses BERT-base \cite{devlin-etal-2019-bert} for representation learning, where it estimates the passage relevance score $\mathbf{D}_k$ for each retrieved passage $d_k$ based on the [CLS] tokens of all retrieved passages $D$, and assigns span relevance scores $S_i$ for each candidate span based on the representations of its start and end tokens.
Finally, the span with the highest span relevance score from the passage with the highest passage relevance score is chosen as the answer.
We refer the readers to \citet{karpukhin2020dense} for more details.

\start{Passage-level Span Voting}
Many extractive QA methods \cite{chen-etal-2017-reading,min2019knowledge,guu2020realm,karpukhin2020dense} measure the probability of span extraction in different retrieved passages independently, despite that their collective signals may provide more evidence in determining the correct answer.
We propose a simple yet effective passage-level span voting mechanism, which aggregates the predictions of the spans in the same surface form from different retrieved passages.
Intuitively, if a text span is considered as the answer multiple times in different passages, it is more likely to be the correct answer.
Specifically,  \ours calculates a normalized score $p(S_i[j])$ for the j-th span in passage $d_i$ during inference as follows: $p (S_i[j]) = \text{softmax} (\mathbf{D})[i] \times \text{softmax} (\mathbf{S_i})[j]$.
\ours then aggregates the scores of the spans with the same surface string among all the retrieved passages as the collective passage-level score.\footnote{We find that the number of spans used for normalization in each passage does not have significant impact on the final performance (we take $N=5$) and using the raw or normalized strings for aggregation also perform similarly.}

\subsection{Generative Reader}
For the generative setup, we use a seq2seq framework where the input is the concatenation of the question and top-retrieved passages and the target output is the desired answer. Such generative readers are adopted in recent methods such as SpanSeqGen~\cite{min2020ambigqa} and Longformer~\cite{beltagy2020longformer}.
Specifically, we use BART-large \cite{lewis2019bart} as the generative reader, which concatenates the question and top-retrieved passages up to its length limit (1,024 tokens, 7.8 passages on average). Generative \ours is directly comparable with SpanSeqGen \cite{min2020ambigqa} that uses the retrieval results of DPR but not comparable with Fusion-in-Decoder (FID) \cite{izacard2020leveraging} since it encodes 100 passages rather than 1,024 tokens and involves more model parameters.

\section{Experiment Setup}

\subsection{Datasets}
We conduct experiments on the open-domain version of two popular QA benchmarks: Natural Questions (NQ) \cite{kwiatkowski-etal-2019-natural} and TriviaQA (Trivia) \cite{joshi-etal-2017-triviaqa}.
The statistics of the datasets are listed in Table~\ref{tab:dataset}.

\begin{table}[ht]
\centering

\resizebox{\columnwidth}{!}{
\scalebox{1}{
\begin{tabular}{llrrr}
\toprule
\textbf{Dataset} & \textbf{Train / Val / Test} & \textbf{Q-len} & \textbf{A-len} & \textbf{\#-A}\\
\midrule
NQ &79,168 / 8,757 / 3,610	&12.5	& 5.2 & 1.2\\
Trivia                 &         78,785 / 8,837 / 11,313  &    20.2  &  5.5 &  13.7\\

\bottomrule
\end{tabular}
}
}
\upv
\caption{Dataset statistics that show the number of samples per data split, the average question (answer) length, and the number of answers for each question.
}
\label{tab:dataset}
\downv
\end{table}

\subsection{Evaluation Metrics}
Following prior studies \cite{karpukhin2020dense}, we use top-k retrieval accuracy to evaluate the performance of the retriever and the Exact Match (EM) score to measure the performance of the reader.

\textit{Top-k retrieval accuracy} is defined as the proportion of questions for which the top-k retrieved passages contain at least one answer span, which is an upper bound of how many questions are ``answerable'' by an extractive reader.

\textit{Exact Match (EM)} is the proportion of the predicted answer spans being exactly the same as (one of) the ground-truth answer(s), after string normalization such as article and punctuation removal.

\subsection{Compared Methods} 
For passage retrieval, we mainly compare with BM25 and DPR, which represent the most used state-of-the-art methods of sparse and dense retrieval for OpenQA, respectively.
For query expansion, we re-emphasize that \ours is the first QE approach designed for OpenQA and most of the recent approaches are not applicable or efficient enough for OpenQA since they have task-specific objectives, require external supervision that was shown to transfer poorly to OpenQA, or take many days to train (Sec.~\ref{sec:related_work}). We thus compare with a classic unsupervised QE method RM3 \cite{abdul2004umass} that does not need external resources for a fair comparison.
For passage reading, we compare with both extractive~\citep{min-etal-2019-discrete,asai2019learning,lee-etal-2019-latent,min2019knowledge,guu2020realm,karpukhin2020dense}  and generative~\citep{brown2020language,roberts2020much,min2020ambigqa,lewis2020retrieval,izacard2020leveraging}  methods when equipping \ours with the corresponding reader.

\begin{table*}[ht]
        \resizebox{1.98\columnwidth}{!}{
        \begin{tabular}{p{17cm}}
            \toprule
            
            \textbf{Question}: when did bat out of hell get released?  \\
             \textbf{Answer}: \colorG{September 1977} \quad \colorB{\{September 1977\}} \\
             \textbf{Sentence}: Bat Out of Hell is the second studio album and the major - label debut by American rock singer Meat Loaf ... released in \colorG{September 1977} on Cleveland International / Epic Records.
             \\ \colorB{\{The album was released in September 1977 on Cleveland International / Epic Records.\}}\\
             \textbf{Title}: \colorG{Bat Out of Hell} \quad \colorB{\{Bat Out of Hell\}}\\
             
            \midrule
                \textbf{Question}: who sings does he love me with reba?  \\
             \textbf{Answer}: \colorR{Brooks \& Dunn}  \quad \colorB{\{Linda Davis\}} \\
             \textbf{Sentence}: \colorG{Linda Kaye Davis} ( born November 26, 1962 ) is an American country music singer. 
             \\ \colorB{\{`` Does He Love You '' is a song written by Sandy Knox and Billy Stritch, and recorded as a duet by American country music artists Reba McEntire and Linda Davis.\}} \\
             \textbf{Title}: \colorG{Does He Love Me} [SEP] \colorG{Does He Love Me (Reba McEntire song)} [SEP] I Do (Reba McEntire album) 
             \colorB{\{Linda Davis [SEP] Greatest Hits Volume Two (Reba McEntire album) [SEP] Does He Love You\}}\\
             
            \midrule
            \textbf{Question}: what is the name of wonder womans mother?  \\
             \textbf{Answer}: \colorR{Mother Magda} \quad\colorB{\{Queen Hippolyta\}} \\
             \textbf{Sentence}: In the Amazonian myths, she is the daughter of the Amazon queen Sifrat and the male dwarf Shuri, and is the mother of Wonder Woman. \quad\colorB{\{Wonder Woman's origin story relates that she was sculpted from clay by her mother Queen Hippolyta and given life by Aphrodite.\}}\\
             \textbf{Title}: \colorG{Wonder Woman} [SEP] \colorG{Diana Prince} [SEP] \colorG{Wonder Woman (2011 TV pilot)} 
             \\ \colorB{\{Wonder Woman [SEP] Orana (comics) [SEP] Wonder Woman (TV series)\}}\\

            \bottomrule
        \end{tabular}
        }
    \upv
    \caption{\textbf{Examples of generated query contexts}. \colorG{Relevant} and \colorR{irrelevant} contexts are shown in green and red. \colorB{Ground-truth references} are shown in the \{braces\}. The issue of generating wrong answers is alleviated by generating other contexts highly related to the question/answer.}
    \label{tab:generation_eg}
    \downv
    \end{table*}

\subsection{Implementation Details}
\start{Retriever}
We use Anserini \cite{yang2017anserini} for text retrieval of BM25 and \ours with its default parameters.
We conduct grid search for the QE baseline RM3 \cite{abdul2004umass}.

\start{Generator}
We use BART-large \cite{lewis2019bart} to generate query contexts in \ours.
When there are multiple desired targets (such as multiple answers or titles), we concatenate them with [SEP] tokens as the reference and remove the [SEP] tokens in the generation-augmented queries.
For Trivia, in particular, we use the value field as the generation target of answer and observe better performance.
We take the checkpoint with the best ROUGE-1 F1 score on the validation set, while observing that the retrieval accuracy of \ours is relatively stable to the checkpoint selection since we do not directly use the generated contexts but treat them as augmentation of queries for retrieval.

\start{Reader}
Extractive \ours uses the reader of DPR with largely the same hyperparameters, which is initialized with BERT-base \cite{devlin-etal-2019-bert} and takes 100 (500) retrieved passages during training (inference).
Generative \ours concatenates the question and top-10 retrieved passages, and takes at most 1,024 tokens as input.
Greedy decoding is adopted for all generation models, which appears to perform similarly to (more expensive) beam search.

\section{Experiment Results}
We evaluate the effectiveness of \ours in three stages: \textit{generation} of query contexts (Sec.~\ref{sec:exp_gen}), \textit{retrieval} of relevant passages (Sec.~\ref{res_retrieval}), and passage \textit{reading} for OpenQA (Sec.~\ref{sec:exp_read}).
Ablation studies are mostly shown on the NQ dataset to understand the drawbacks of \ours since it achieves better performance on Trivia.

\subsection{Query Context Generation}
\label{sec:exp_gen}

\start{Automatic Evaluation}
To evaluate the quality of the generated query contexts, we first measure their lexical overlap with the ground-truth query contexts.
As suggested by the nontrivial ROUGE scores in Table~\ref{tab:generation_num}, \ours does learn to generate meaningful query contexts that could help the retrieval stage.
We next measure the lexical overlap between the query and the ground-truth passage.
The ROUGE-1/2/L F1 scores between the original query and ground-truth passage are 6.00/2.36/5.01, and those for the generation-augmented query are 7.05/2.84/5.62 (answer), 13.21/6.99/10.27 (sentence), 7.13/2.85/5.76 (title) on NQ, respectively. Such results further demonstrate that the generated query contexts significantly increase the word overlap between the queries and the positive passages, and thus are likely to improve retrieval results.\footnote{We use F1 instead of recall to avoid the unfair favor of (longer) generation-augmented query.}

\begin{table}[ht]
\centering

\scalebox{.8}{
\begin{tabular}{lrrr}
\toprule
\textbf{Context}  & \textbf{ROUGE-1} & \textbf{ROUGE-2} & \textbf{ROUGE-L}\\
\midrule
Answer	&33.51	& 20.54 & 33.30\\
Sentence         &    37.14  &  24.71 &  33.91\\
Title         &    43.20  &  32.11 &  39.67\\

\bottomrule
\end{tabular}
}

\upv
\caption{\textbf{ROUGE F1 scores of the generated query contexts} on the validation set of the NQ dataset.
}
\label{tab:generation_num}
\downv
\end{table}

\begin{table*}[ht]
\centering

\resizebox{2\columnwidth}{!}{
\begin{tabular}{l ccccc | ccccc}
  \toprule
    \multirow{2}{*}{ \textbf{Method}}  & \multicolumn{5}{c|}{\textbf{NQ}} & \multicolumn{5}{c}{\textbf{Trivia}} \\
    & Top-5 & Top-20 & Top-100 & Top-500 & Top-1000 & Top-5 & Top-20 & Top-100 & Top-500 & Top-1000 \\
\midrule
    BM25 (ours)        & 43.6 & 62.9& 78.1 & 85.5 & 87.8 & 67.7 & 77.3 & 83.9 & 87.9 & 88.9  \\
    BM25 +RM3        & 44.6 & 64.2& 79.6 & 86.8 & 88.9 & 67.0 & 77.1 & 83.8 & 87.7 & 88.9  \\
    DPR         & \underline{68.3} & \underline{80.1} & \underline{86.1} & 90.3 & 91.2 & 72.7 & 80.2 & 84.8 & - & - \\
    \ours       &60.9 & 74.4 & 85.3 & \underline{90.3} & \underline{91.7} & \underline{73.1} & \underline{80.4} & \underline{85.7} & \textbf{88.9} & \textbf{89.7} \\
    \oursPlus  &\textbf{70.7} & \textbf{81.6} & \textbf{88.9} & \textbf{92.0} & \textbf{93.2} & \textbf{76.0} & \textbf{82.1} & \textbf{86.6} & - & - \\
  
  \bottomrule
\end{tabular}
}

\caption[Caption]{\textbf{Top-k retrieval accuracy on the test sets}. The baselines are evaluated by ourselves and better than reported in \citet{karpukhin2020dense}. \ours helps BM25 to achieve comparable or better performance than DPR. Best and second best methods are \textbf{bold} and \underline{underlined}, respectively.}
\label{tab:top_k_acc}
\end{table*}

\start{Case Studies}
In Table~\ref{tab:generation_eg}, we show several examples of the generated query contexts and their ground-truth references.
In the first example, the correct album release date appears in both the generated answer and the generated sentence, and the generated title is the same as the Wikipedia page title of the album.
In the last two examples, the generated answers are wrong but fortunately, the generated sentences contain the correct answer and (or) other relevant information and the generated titles are highly related to the question as well, which shows that different query contexts are complementary to each other and the noise during query context generation is thus reduced.

\subsection{Generation-Augmented Retrieval}
\label{res_retrieval}

\start{Comparison w. the state-of-the-art}
We next evaluate the effectiveness of \ours for retrieval.
In Table~\ref{tab:top_k_acc}, we show the top-k retrieval accuracy of BM25, BM25 with query expansion (+RM3) \cite{abdul2004umass}, DPR~\citep{karpukhin2020dense}, \ours, and \oursPlus (\ours+DPR).

On the NQ dataset, while BM25 clearly underperforms DPR regardless of the number of retrieved passages, the gap between \ours and DPR is significantly smaller and negligible  when $k \geq 100$. When $k \geq 500$, \ours is slightly better than DPR despite that it simply uses BM25 for retrieval.
In contrast, the classic QE method RM3, while showing marginal improvement over the vanilla BM25, does not achieve comparable performance with \ours or DPR.
By fusing the results of \ours and DPR in the same way as described in Sec.~\ref{sec:retrieval}, we further obtain consistently higher performance than both methods, with top-100 accuracy 88.9\% and top-1000 accuracy 93.2\%.

On the Trivia dataset, the results are even more encouraging -- \ours achieves consistently better retrieval accuracy than DPR when $k \geq 5$. On the other hand, the difference between BM25 and BM25 +RM3 is negligible, which suggests that naively considering top-ranked passages as relevant (\ie, pseudo relevance feedback) for QE does not always work for OpenQA.
Results on more cutoffs of $k$ can be found in App.~\ref{sec_app:top_k_acc}.

\begin{figure}[ht]
    \centering
    \includegraphics[width=0.99\linewidth]{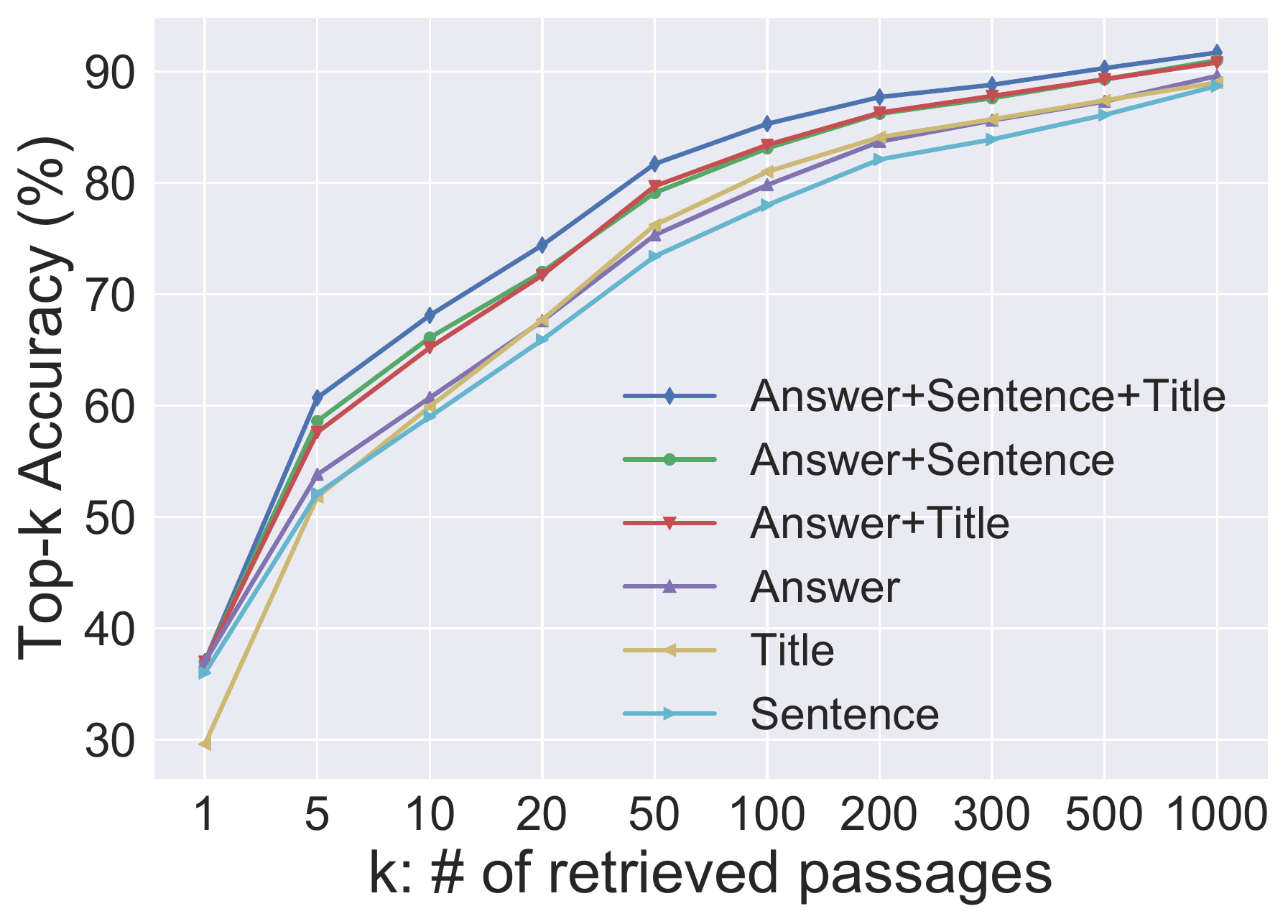}
    \upv
    \upvf
    \vspace{-.2cm}
    \caption{\textbf{Top-k retrieval accuracy} on the test set of NQ when fusing retrieval results of different generation-augmented queries.}
    \label{fig:fuse}
    \downv
\end{figure}

\start{Effectiveness of diverse query contexts}
In Fig.~\ref{fig:fuse}, we show the performance of \ours when different query contexts are used to augment the queries.
Although the individual performance when using each query context is somewhat similar, fusing their retrieved passages consistently leads to better performance, confirming that different generation-augmented queries are complementary to each other (recall examples in Table~\ref{tab:generation_eg}).

\start{Performance breakdown by question type}
In Table~\ref{tab:breakdown}, we show the top-100 accuracy of the compared retrieval methods per question type on the NQ test set. Again, \ours outperforms BM25 on all types of questions significantly and \oursPlus achieves the best performance across the board, which further verifies the effectiveness of \ours.

\begin{table}[ht]
\centering

\resizebox{1.\columnwidth}{!}{
\begin{tabular}{l rcccc }
  \toprule
     \textbf{Type} & \textbf{Percentage} & \textbf{BM25} & \textbf{DPR} & \textbf{\ours}  & \textbf{\oursPlus}  \\
\midrule
Who 	& 	37.5\%& 82.1 & \underline{88.0} & 87.5 & \textbf{90.8} \\
When 	& 	19.0\%& 73.1 & \underline{86.9} & 83.8 & \textbf{88.6} \\
What 	& 	15.0\%& 76.5 & \underline{82.6} & 81.5 & \textbf{86.0} \\
Where 	& 	10.9\%& 77.4 & \underline{89.1} & 87.0 & \textbf{90.8} \\
Other 	& 	9.1\%& 79.3 & 78.1 & \underline{81.8} & \textbf{84.2} \\
How 	& 	5.0\%& 78.2 & \underline{83.8} & 83.2 & \textbf{85.5} \\
Which 	& 	3.3\%& 89.0 & 90.7 & \underline{94.1} & \textbf{94.9} \\
Why 	& 	0.3\%& 90.0 & 90.0 & 90.0 & 90.0 \\

  \bottomrule
\end{tabular}
}

\caption[Caption]{\textbf{Top-100 retrieval accuracy breakdown of question type on NQ}. Best and second best methods in each category are \textbf{bold} and \underline{underlined}, respectively.}
\label{tab:breakdown}
\end{table}

\begin{table}[t]
\centering

\resizebox{1.05\columnwidth}{!}{
\begin{tabular}{clcccc}
  \cmidrule[0.06em]{2-5}
    &\textbf{Method} & \textbf{NQ} & \multicolumn{2}{c}{\textbf{Trivia}} \\

\cmidrule{2-5}

  \multirow{9}{*}{ \rotatebox[origin=c]{90}{Extractive}} &Hard EM~\citep{min-etal-2019-discrete}          & 28.1 & 50.9    & - \\
  &Path Retriever~\citep{asai2019learning}  & 32.6 & -    & -    \\
  &ORQA~\citep{lee-etal-2019-latent}               & 33.3 & 45.0 & -    \\
  &Graph Retriever~\citep{min2019knowledge} & 34.5 & 56.0 & -    \\
  &REALM~\citep{guu2020realm}               & 40.4 & -    & -    &  \\
  &DPR~\citep{karpukhin2020dense}           & 41.5 & 57.9 & -   \\
  &BM25 (ours) & 37.7 & 60.1 & -& \\
  &\ours & \textbf{41.8} & \textbf{62.7} & \textbf{74.8} & \\
  &\oursPlus & \textbf{43.8}& - & - & \\
  
\cmidrule{2-5}

\multirow{8}{*}{ \rotatebox[origin=c]{90}{Generative}} &GPT-3~\citep{brown2020language}    & 29.9 & - & 71.2  \\
    &T5~\citep{roberts2020much} & 36.6 & 60.5 & - \\
  
  &SpanSeqGen~\citep{min2020ambigqa}        & 42.2 & -    & -    \\
  &RAG~\citep{lewis2020retrieval}           & 44.5 & 56.1 & 68.0  \\
  &FID \cite{izacard2020leveraging}   & \textbf{51.4} & \textbf{67.6} & \textbf{80.1}  \\
  &BM25 (ours) & 35.3 & 58.6 & -& \\
  &\ours & 38.1 & \textbf{62.2} & - & \\
  &\oursPlus & \textbf{45.3} & - & - & \\

  \cmidrule[0.06em]{2-5}
\end{tabular}
}

\caption[Caption]{\textbf{End-to-end comparison with the state-of-the-art methods in EM}.
For Trivia, the left column denotes the open-domain test set and the right is the hidden Wikipedia test set on the public leaderboard.}
\label{tab:sota}
\end{table}

\subsection{Passage Reading with \ours}
\label{sec:exp_read}

\start{Comparison w. the state-of-the-art}
We show the comparison of end-to-end QA performance of extractive and generative methods in Table~\ref{tab:sota}.
Extractive \ours achieves state-of-the-art performance among extractive methods on both NQ and Trivia datasets, despite that it is more lightweight and computationally efficient.
Generative \ours outperforms most of the generative methods on Trivia but does not perform as well on NQ, which is somewhat expected and consistent with the performance at the retrieval stage, as the generative reader only takes a few passages as input and \ours does not outperform dense retrieval methods on NQ when $k$ is very small.
However, combining \ours with DPR achieves significantly better performance than both methods or baselines that use DPR as input such as SpanSeqGen~\citep{min2020ambigqa} and RAG~\citep{lewis2020retrieval}.
Also, \ours outperforms BM25 significantly under both extractive and generative setups, which again shows the effectiveness of the generated query contexts, even if they are heuristically discovered without any external supervision.

The best performing generative method FID \cite{izacard2020leveraging} is not directly comparable as it takes more (100) passages as input. As an indirect comparison, \ours performs better than FID when FID encodes 10 passages (cf. Fig.~2 in \citet{izacard2020leveraging}).
Moreover, since FID relies on the retrieval results of DPR as well, we believe that it is a low-hanging fruit to replace its input with \ours or \oursPlus and further boost the performance.\footnote{This claim is later verified by the best systems in the NeurIPS 2020 EfficientQA competition \cite{min2021neurips}.}
We also observe that, perhaps surprisingly, extractive BM25 performs reasonably well, especially on the Trivia dataset, outperforming many recent state-of-the-art methods.\footnote{We find that taking 500 passages during reader inference instead of 100 as in \citet{karpukhin2020dense} improves the performance of BM25 but not DPR.} Generative BM25 also performs competitively in our experiments.

\start{Model Generalizability}
Recent studies \cite{lewis2020question} show that there are significant question and answer overlaps between the training and test sets of popular OpenQA datasets.
Specifically, 60\% to 70\% test-time answers also appear in the training set and roughly 30\% test-set questions have a near-duplicate paraphrase in the training set. Such observations suggest that many questions might have been answered by simple question or answer memorization.
To further examine model generalizability, we study the per-category performance of different methods using the annotations in \citet{lewis2020question}.

\begin{table}[ht]
\centering

\resizebox{\columnwidth}{!}{
\scalebox{1}{
\begin{tabular}{lcccc}
\toprule
\textbf{Method} & \textbf{Total} & \multicolumn{1}{m{1.5cm}}{\centering \textbf{Question Overlap}} & \multicolumn{1}{m{1.5cm}}{\centering \textbf{Answer Overlap Only}} & \multicolumn{1}{m{1.5cm}}{\centering \textbf{No Overlap}}\\
\midrule

DPR & 41.3  & \textbf{69.4} &  34.6 &  19.3\\
\oursPlus (E) &\textbf{43.8}	&66.7	&\textbf{38.1}	&\textbf{23.9} \\
\midrule
BART & 26.5	& 67.6	& 10.2 & 0.8\\
RAG & 44.5	& \textbf{70.7}	& 34.9 & 24.8\\
\oursPlus (G) &\textbf{45.3}	&67.9	&\textbf{38.1}	&\textbf{27.0} \\

\bottomrule
\end{tabular}
}
}
\upv
\caption{\textbf{EM scores with question-answer overlap category breakdown  on NQ.} (E) and (G) denote extractive and generative readers, respectively. Results of baseline methods are taken from \citet{lewis2020question}. The observations on Trivia are similar and omitted.
}
\label{tab:overlap}
\downv
\end{table}

As listed in Table~\ref{tab:overlap}, for the \textit{No Overlap} category, \oursPlus (E) outperforms DPR on the extractive setup and \oursPlus (G) outperforms RAG on the generative setup, which indicates that better end-to-end model generalizability can be achieved by adding \ours for retrieval.
\oursPlus also achieves the best EM under the \textit{Answer Overlap Only} category.
In addition, we observe that a closed-book BART model that only takes the question as input performs much worse than additionally taking top-retrieved passages, \ie, \oursPlus (G), especially on the questions that require generalizability.
Notably, all methods perform significantly better on the \textit{Question Overlap} category, which suggests that the high \textit{Total} EM is mostly contributed by question memorization. That said, \oursPlus appears to be less dependent on question memorization given its lower EM for this category.\footnote{The same ablation study is also conducted on the retrieval stage and similar results are observed. More detailed discussions  can be found in App.~\ref{sec_app:top_k_acc}.}

\subsection{Efficiency of \ours }
\label{sec:runtime}

\ours is efficient and scalable since it uses sparse representations for retrieval and does not involve time-consuming training process such as RL \cite{nogueira-cho-2017-task,liu2019generative}. The only overhead of \ours is on the generation of query contexts and the retrieval with generation-augmented (thus longer) queries, whose computational complexity is significantly lower than other methods with comparable retrieval accuracy.

\begin{table}[t]
\centering

\resizebox{\columnwidth}{!}{
\scalebox{1}{
\begin{tabular}{lccc}
\toprule
\textbf{} & \textbf{Training} & \textbf{Indexing} & \textbf{Retrieval} \\
\midrule
DPR & 24h w. 8 GPUs	& 17.3h w. 8 GPUs	&  30 min w. 1 GPU \\
\ours   &   3 $\sim$ 6h w. 1 GPU       &    0.5h w. 35 CPUs  & 5 min w. 35 CPUs  \\
\bottomrule
\end{tabular}
}
}
\upv
\caption{\textbf{Comparison of computational cost between DPR and \ours at different stages.} The training time of \ours is for one generation target but  different generators can be trained in parallel.
}
\label{tab:runtime}
\downv
\end{table}

We use Nvidia V100 GPUs and Intel Xeon Platinum 8168 CPUs in our  experiments.
As listed in Table~\ref{tab:runtime}, the training time of \ours is 3 to 6 hours on 1 GPU depending on the generation target.
As a comparison, REALM \cite{guu2020realm} uses 64 TPUs to train for 200k steps during pre-training alone and DPR \cite{karpukhin2020dense} takes about 24 hours to train with 8 GPUs.
To build the indices of Wikipedia passages, \ours only takes around 30 min with 35 CPUs, while DPR takes 8.8 hours on 8 GPUs to generate dense representations and another 8.5 hours to build the FAISS index \cite{JDH17}.
For retrieval, \ours takes about 1 min to generate one query context with 1 GPU, 1 min to retrieve 1,000 passages for the NQ test set with answer/title-augmented queries and 2 min with sentence-augmented queries using 35 CPUs. In contrast, DPR takes about 30 min on 1 GPU.

\section{Conclusion}
In this work, we propose Generation-Augmented Retrieval and demonstrate that the relevant contexts generated by PLMs without external supervision can significantly enrich query semantics and improve retrieval accuracy.
Remarkably, \ours with sparse representations performs similarly or better than state-of-the-art methods based on the dense representations of the original queries. \ours can also be easily combined with dense representations to produce even better results.
Furthermore, \ours achieves state-of-the-art end-to-end performance on extractive OpenQA and competitive performance under the generative setup.

\section{Future Extensions}
\start{Potential improvements}
There is still much space to explore and improve for \ours in future work.
For query context generation, one can explore multi-task learning to further reduce computational cost and examine whether different contexts can mutually enhance each other when generated by the same generator.
One may also sample multiple contexts instead of greedy decoding to enrich a query.
For retrieval, one can adopt more advanced fusion techniques based on both the ranking and score of the passages.
As the generator and retriever are largely independent now, it is also interesting to study how to jointly or iteratively optimize generation and retrieval such that the generator is aware of the retriever and generates query contexts more beneficial for the retrieval stage.
Last but not least, it is very likely that better results can be obtained by more extensive hyper-parameter tuning.

\start{Applicability to other tasks}
Beyond OpenQA, \ours also has great potentials for other tasks that involve text matching such as conversation utterance selection \cite{lowe2015ubuntu,dinan2020second} or information retrieval \cite{nguyen2016ms,craswell2020overview}.
The default generation target is always available for supervised tasks. For example, for conversation utterance selection one can use the reference utterance as the default target and then match the concatenation of the conversation history and the generated utterance with the provided utterance candidates.
For article search, the default target could be (part of) the ground-truth article itself.
Other generation targets are more task-specific and can be designed as long as they can be fetched from the latent knowledge inside PLMs and are helpful for further text retrieval (matching).
Note that by augmenting (expanding) the queries with heuristically discovered relevant contexts extracted from PLMs instead of reformulating them, \ours bypasses the need for external supervision to form the original-reformulated query pairs.

\section*{Acknowledgments}
We thank Vladimir Karpukhin, Sewon Min, Gautier Izacard, Wenda Qiu, Revanth Reddy, and Hao Cheng for helpful discussions. We thank the anonymous reviewers for valuable comments.

\bibliographystyle{acl_natbib}
\bibliography{anthology,main}

\newpage
\clearpage
\appendix

\section{More Analysis of Retrieval Performance}
\label{sec_app:top_k_acc}
\begin{figure}[ht]
    \centering
    \includegraphics[width=0.99\linewidth]{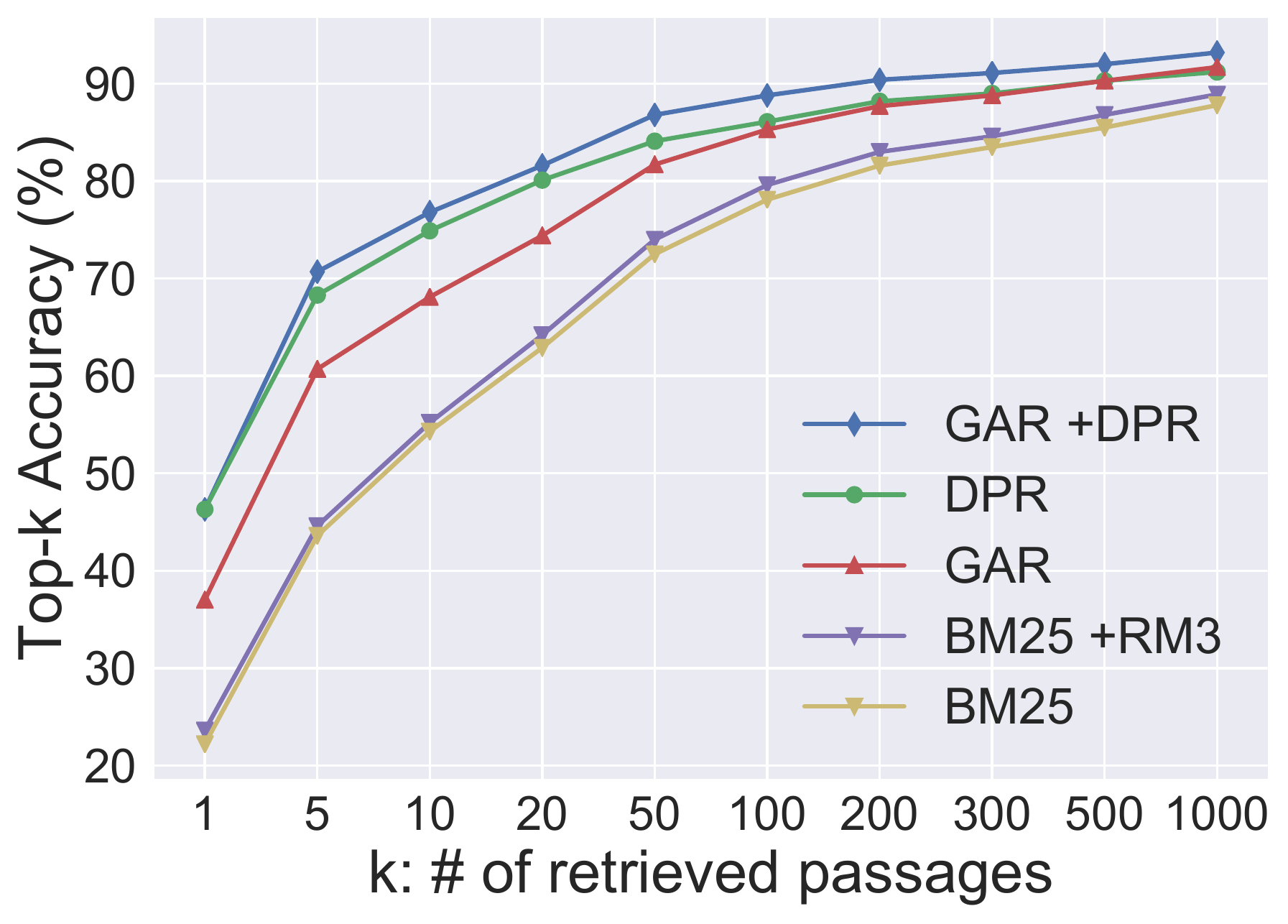}
    \upv
    \upvf
    \vspace{-.4cm}
    \caption{\textbf{Top-k retrieval accuracy of sparse and dense methods on the test set of NQ.} \ours improves BM25 and achieves comparable or  better performance than DPR  when $k \geq 100$.}
    \label{fig:top_k_acc}
    \downv
\end{figure}

\begin{figure}[ht]
    \centering
    \includegraphics[width=0.99\linewidth]{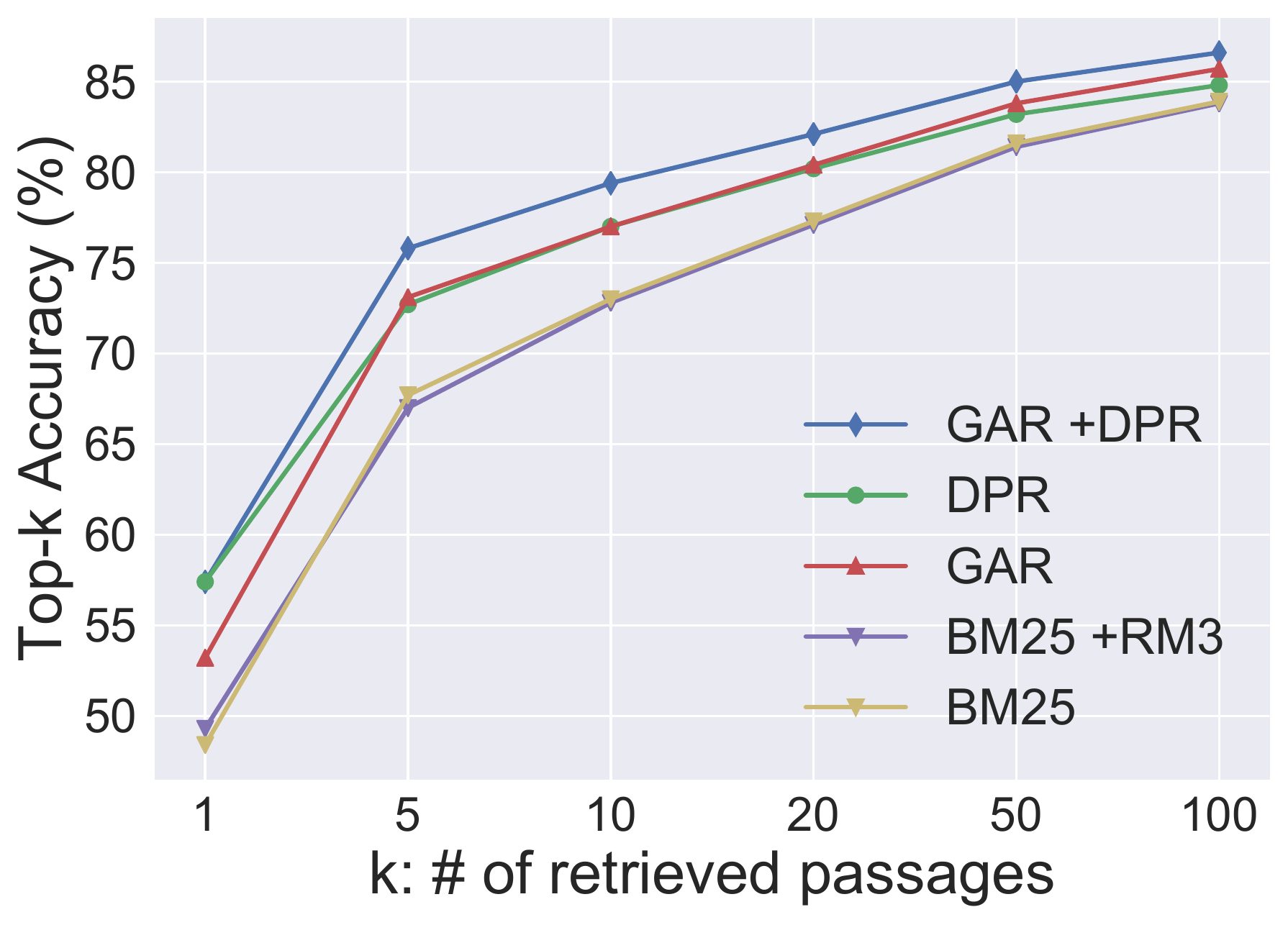}
    \upv
    \upvf
    \vspace{-.4cm}
    \caption{\textbf{Top-k retrieval accuracy on the Trivia test set.} \ours achieves better results than DPR when $k \geq 5$.}
    \label{fig:top_k_acc_trivia}
    \downv
\end{figure}

We show the detailed results of top-k retrieval accuracy of the compared methods in Figs.~\ref{fig:top_k_acc} and~\ref{fig:top_k_acc_trivia}.
\ours performs comparably or better than DPR when $k \geq 100$ on NQ and $k \geq 5$ on Trivia.

We show in Table~\ref{tab:overlap_retrieval} the retrieval accuracy breakdown using the question-answer overlap categories. The most significant gap between BM25 and other methods is on the \textit{Question Overlap} category, which coincides with the fact that BM25 is unable to conduct question paraphrasing (semantic matching). \ours helps BM25 to bridge the gap by providing the query contexts and even outperform DPR in this category.  Moreover, \ours consistently improves over BM25 on other categories and \oursPlus outperforms DPR as well.
\begin{table}[ht]
\centering

\resizebox{\columnwidth}{!}{
\scalebox{1}{
\begin{tabular}{lcccc}
\toprule
\textbf{Method} & \textbf{Total} & \multicolumn{1}{m{1.5cm}}{\centering \textbf{Question Overlap}} & \multicolumn{1}{m{1.5cm}}{\centering \textbf{Answer Overlap Only}} & \multicolumn{1}{m{1.5cm}}{\centering \textbf{No Overlap}}\\
\midrule

BM25 & 78.8 & 81.2 & 85.1 & 70.6 \\
DPR & 86.1 & 93.2 & 89.5 & 76.8 \\
\ours & 85.3 & 94.1 & 87.9 & 73.7 \\
\oursPlus &\textbf{88.9} & \textbf{96.3} & \textbf{91.7} & \textbf{79.8} \\

\bottomrule
\end{tabular}
}
}
\upv
\caption{\textbf{Top-100 retrieval accuracy by question-answer overlap categories on the NQ test set.} 
}
\label{tab:overlap_retrieval}
\downv
\end{table}

\end{document}